\documentclass{llncs}
\usepackage{makeidx}  % allows for indexgeneration
\usepackage{epsfig}
\usepackage{gb4e}
\usepackage{algorithm}
\usepackage{algorithmic}

\title{On Introspection, Metacognitive Control and Augmented Data Mining Live Cycles}
\author{Daniel Sonntag}
\institute{German Research Center for Artificial Intelligence\\ 
 66123 Saarbr{\"u}cken, Germany \\
 \email{sonntag@dfki.de}}

\begin{document}
\maketitle

\begin{abstract}
We discuss metacognitive modelling as an enhancement to cognitive
modelling and computing. Metacognitive control mechanisms should
enable AI systems to self-reflect, reason about their
actions, and to adapt to new situations. In this respect, we propose
implementation details of a knowledge taxonomy and an augmented data
mining life cycle which supports a live integration of obtained models.\\\\
{\bf Keywords:} Metacognitive Modelling, Data Mining
\end{abstract}

\section{Introduction}

Cognitive computing is the development of computer techniques to
emulate human perception, intelligence, and problem solving.
Cognitive models are equipped with artificial sensors and actuators
which are integrated and embedded into physical systems or ambient
intelligence environments to act in the physical world. The goal is to
have cognitive capabilities and to perform cognitive control (e.g.,
see \cite{beetz07}).  To overcome problems in shared control (of,
e.g., navigating robots \cite{rossSVKB04}), direct communication (in
natural language dialogue) between a human participant and a technical
control architecture can be employed. This could be used for mutual
disambiguation of multiple sensory modalities in a learning
environment.  As one of the major topics of sensory-based control
mechanisms, automatic perception learning by introspection and
relevance feedback could help in this disambiguation task. In order to
pursue the idea of cognitive systems able to self-reflect, reason
about their actions, and to adapt to new situations, metacognitive
strategies can be employed.

In this paper, we will present the core idea of a metacognitive control
model of machine learning with respect to problem solving capabilities
to be exemplified by improving autonomous reaction behaviour.

We start by clarifying the term {\it metacognition}.  {\it
  Metacognition} is cognition about cognition.  It can, in principle,
enable artificial intelligence systems to monitor and control
themselves, choose goals, assess progress, and adopt new strategies
for achieving goals.\footnote{For example, students preparing for an
  exam judge about the relative difficulty of the learning material
  and use this for study strategies. The resulting reasoning task is a
  second-order reasoning process about the own learning abilities
  called meta-reasoning or, more generally, metacognition.}
\cite{derry89} associates metacognitive components with the ability of
a subject (or an intelligent agent in general) to orchestrate and
monitor knowledge of the problem solving process; \cite{davidson94}
argues that metacognitive abilities correlate with standard measures
of intelligence; \cite{brachman02} talks about systems that know what
they are doing.

 Here, we adopt the growing interest in metacognitive
 strategies\footnote{IBM Autonomic Computing Initiative, {\it
     http://www.research.ibm.com/autonomic/},\\and, e.g., DARPA
   Information Processing Technology Office on Cognitive Systems,
   \\ {\it http://www.darpa.mil/ipto/thrust\_areas/thrust\_cs.asp}.}
 for AI systems to build a metacognitive model for adaptable AI
 systems, which involves computational models of self-representation
 and self-awareness. Ontologies represent the knowledge groundwork for
 the self-representation of a system information state to be included
 into a metacognitive model.\footnote{\cite{mccarthy68} outlines that
   for intelligent behaviour, a declarative knowledge model must be
   created first. Examination of, e.g., own beliefs would then be
   possible when the beliefs are explicitly represented.  McCarthy
   sees introspection as essential for human level intelligence (and
   not a mere epiphenomenon) \cite{mccarthy95}.} For example, McCarthy
 defines the term {\it introspection} as a machine having a belief
 about its own mental state rather than a belief about propositions
 concerning the world.

According to this explanation of metacognition we hypothesise that
researchers in adaptable AI systems should investigate in
metacognition because it can help us:

\begin{enumerate}
\item address the difficulty to write down control management
  rules. Rules may not be obvious, tangible, or identifiable, or they
  may present an engineering overhead.
\item provide self-improvement through adaptation and customisation.
\item offer designs for never-ending learning.
\item integrate a variety of previously isolated findings: dialogue
  architectures, finite state strategies, information states,
  (un)supervised learning, stacked generalisation, reinforcement
  learning, interactive learning, and embedded data mining.
\end{enumerate} 

Apart from its complexity, metacognition highlights an empirically
tractable model creation and verification process.

\section{Model, Introspective View and Control}

We use the term {\it model} in the sense given by \cite{minsky68}:

\begin{quote}
To an observer $B$, an object $A^\bullet$ is a model of an object $A$
to the extent that $B$ can use $A^\bullet$ to answer questions that
interest him about $A$.
\end{quote}

$A$ can be the world or a specific sub-domain such as the football
domain. To answer questions about the football domain, an $A^\bullet$
has to be constructed. $A^\bullet$ corresponds to an ontological
knowledge base which contains facts about the sub-domain and the
knowledge how to communicate the facts. This level of knowledge
representation is basically implemented by state-of-the-art semantic
technologies.  Intelligent interaction systems for dialogical
interaction with the Semantic Web (e.g., SmartWeb
\cite{sonntagetal07}) can be built on top of this representation of
domain knowledge (e.g., dialogue and football knowledge).

Contemporary AI introduces the notion of ontologies as a knowledge
representation mechanism (e.g., see \cite{fenseletal:03}) for the
operational AI models we are interested in. The object level
represents the world and the domain of interest; in addition, the domain
ontologies should contain mental concepts about communication and
control structures; and for processing user feedback, a representation
of natural communication (natural language dialogue) is required. When these
concepts can be used to maintain an information state, a model of
introspection can be derived from it. Then, self-reflective knowledge
can be provided by the introspective AI system management facility
which holds an introspective view of the object level. More precisely,
an {\it introspective view} is obtained from introspective reports,
i.e., interpretations of data records of process data as a description
of the internal processes under observation.  In this respect, we
recognise introspection in the same way as done by
\cite{nelsonnarens90}:

\begin{quote}
We view introspective reports as data to be explained, in contrast to
the Structuralists' view of introspective reports as descriptions of
internal processes; i.e., we regard introspection not as a conduit to
the mind but rather as a source of data to be accounted for by
postulated internal processes.\footnote{This important quote basically
  states that metacognition as proposed here is not a reconstruction
  of the respective human intelligence apparatus---in accord with
  technical cognitive AI system research.}
\end{quote}

Thus, the introspective view can be implemented by the output of a
meta-level data generalisation process while reporting on the
object-level behaviour. Metadata providers decide which kind of
information is to be included in the introspective reports.  On the
meta-level, meta-models can be generated with the help of machine
learning and data mining algorithms. A knowledge taxonomy helps
differentiate between the different knowledge levels, especially the
knowledge levels obtained from the machine learning experiments.

\subsection{Meta Knowledge Taxonomy}

\begin{figure}
\begin{center}
\epsfig{file=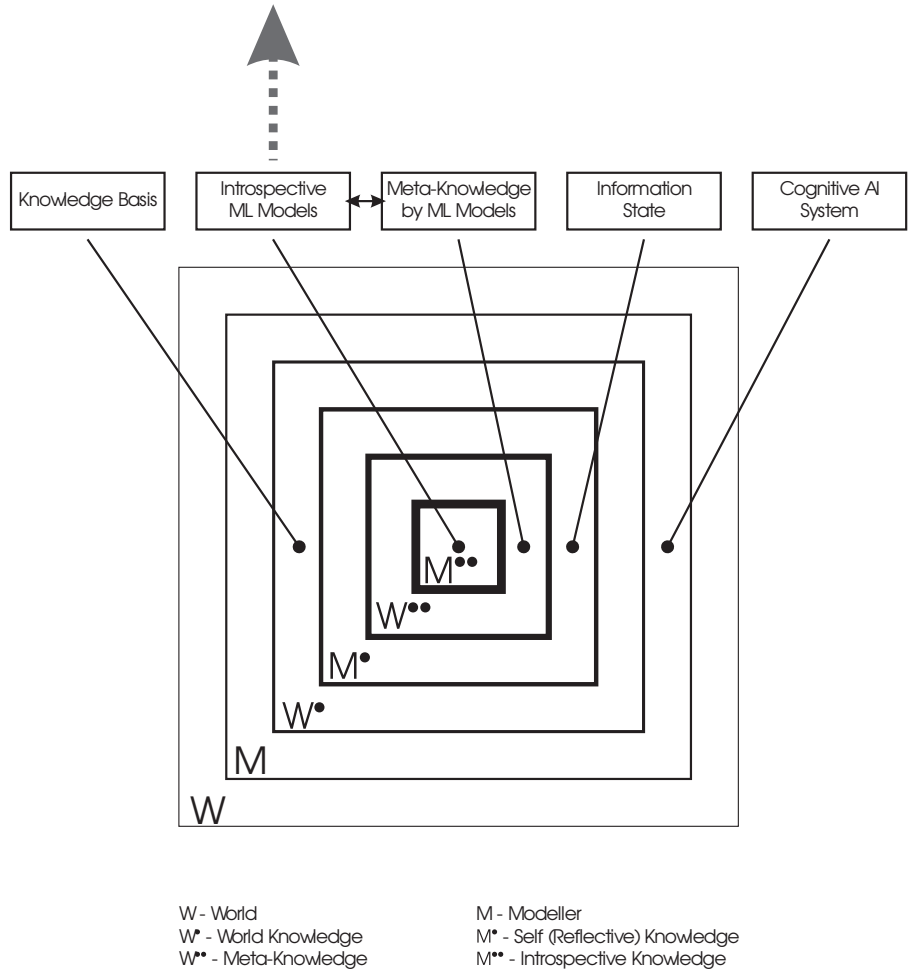, width=1.0\textwidth} 
\end{center}
\caption{Meta Knowledge Taxonomy for metacognition}
\label{ktaxonomy}
\end{figure}

In order to integrate learning schemes---i.e. to learn meta-level
action strategies from experience---we propose a meta knowledge
taxonomy (figure \ref{ktaxonomy}). Consider a world ($W$) and a
modeller ($M$) who exists in the world, and who can be a human or an
intelligent computer agent. A knowledge taxonomy can be constructed to
include the modelling of the world and the modeller (according to some
articles in \cite{minsky86}). In this paper, we provide the
implementations of this knowledge taxonomy by using semantic
technologies and machine learning.

The model of the world ($W^\bullet$) is used to answer questions about
$W$. In order to answer questions about the modeller himself, we
introduce a model of the modeller ($M^\bullet$). $M^\bullet$ is the
self (reflective) knowledge of the agent in the world. Before
explaining the crucial layers of introspective knowledge for complex
AI systems, we should explain our idea of how to map and implement the
first four layers. The world $W$ is the application we have in mind,
with the capability to adapt to new environmental conditions. Thereby,
the processing system is the modeller $M$. Accordingly, $W^\bullet$ is
the processing system's knowledge basis, the ontology terminology box
of the AI system's application domain.  $M^\bullet$ is the internal
state of the dialogue system which we implement as information state,
consisting of assertion box instances, according to the ontology. It
represents the self knowledge of the system in the running state. If
the information state contains information about the system itself,
the modeller's self knowledge can be called reflective.\\

$W^{\bullet\bullet}$ is the model of the world knowledge; it contains
the meta knowledge in order to reason about the questions concerning
the world knowledge. Some typical questions for ontological knowledge
bases are whether the classes and relations adequately describe the
application domain, and whether the descriptive representation of
domain processes provides a motivated conclusive representation of the
situation in terms of {\it content-describing features}. We implement
this meta knowledge layer with machine learning models: if the models
created by the attributes derived from ontology instances have
positive evaluation characteristics (for example, high cross-validated
classification accuracy or reasonable symbolic association rules or
decision trees), we adequately describe the world knowledge by meta
knowledge. (A task-based evaluation of ontologies in a specific
application domain is meta knowledge, too.) $M^{\bullet\bullet}$ is
the knowledge that can be extracted from the processing system while
running the system in the current environment. This self-reflective
knowledge can be used to adapt to other processing strategies, for
example control signals, if the current one fails. At this layer, we
are able to recognise how all other knowledge layers work together to
performing a particular task in the AI system's application domain.\\

Both $W^{\bullet\bullet}$ and $M^{\bullet\bullet}$ can be used to
build {\it decision-oriented operationalised management rules}.
Decision-oriented means that any of the reaction duties are directly
triggered or effected.  Operationalised rules means that the control
rules derived from the machine learning models are in a directly
executable format (e.g., association rules) or can be translated into
these. The operationalisation itself can be undertaken manually or
automatically.  To sum up the intention of the meta knowledge taxonomy
for metacognition. The modelling by a knowledge taxonomy provides
abstract solution for the problem of how
\begin{itemize}
\item to monitor system performance;
\item to adapt a problem solving strategy according to performance classification;
\item to build operational machine learning models.
\end{itemize}

\section{Augmented Data Mining Live Cycle}

\begin{figure}
\centering \includegraphics[width=0.9\textwidth]{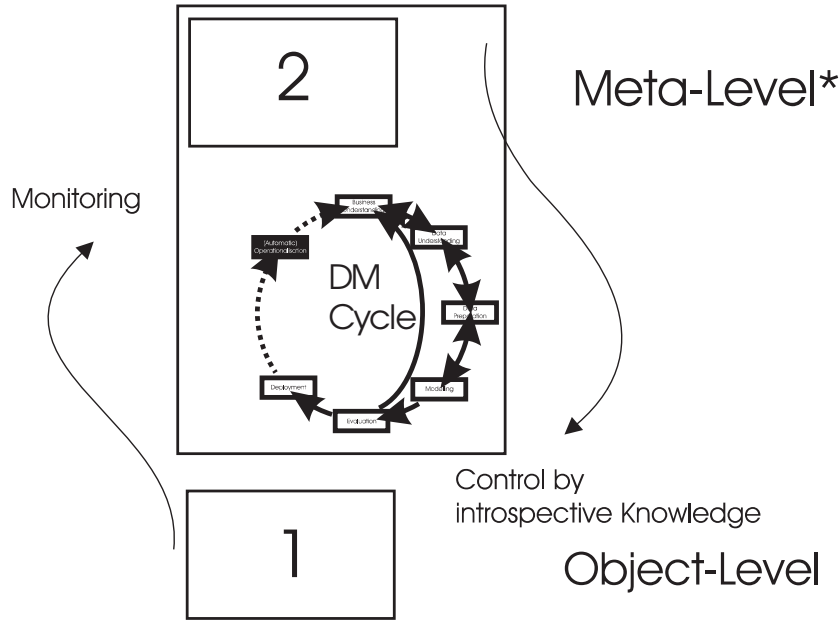}
\caption[Meta-level Control]{The meta-level control is established by
  the embedding of introspective knowledge for control. The augmented
  data mining cycle is shown in figure \ref{cycle}.}
\label{crispy}
\end{figure}

The implementations of the knowledge taxonomy are given by the
processing system, the (ontological) knowledge basis, the information
state, the meta knowledge by ML models, and the introspective ML
models. Thereby, the theory  combines top-down approaches (i.e.,
ontological knowledge representation) with bottom-up approaches (i.e.,
empirical process data model exploitation).  The later means information state
features aggregation and data mining  by combining declarative and
procedural knowledge.  {\it Metacognitive control} is the application
of the introspective knowledge gained on the meta-level by controlling 
the object-level, as illustrated in figure \ref{crispy}. According to
control theory, we are not only able to vary parameters of the object
level control in real-time, but augment the object-level (cognitive)
reasoning process by learned meta-models.  Hence, the {\it
  metacognitive control} idea includes planning, monitoring,
authoring, integration, and evaluation.

\begin{figure}
\begin{center}
\epsfig{file=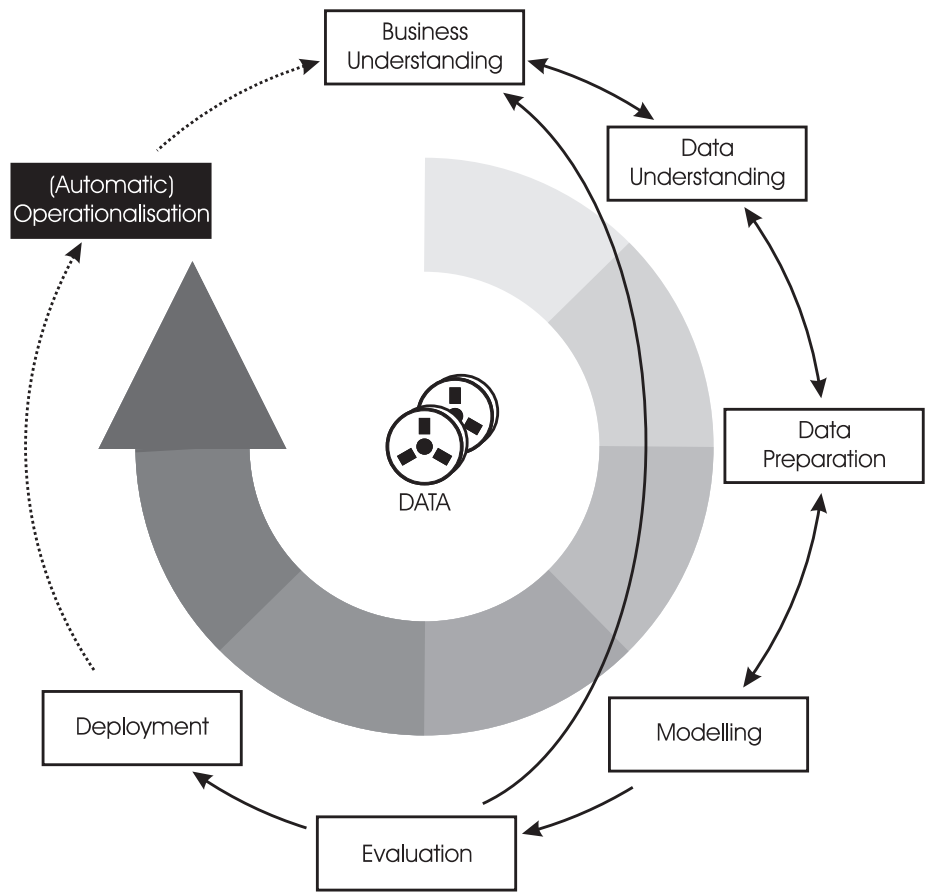, width=1.0\textwidth}
\end{center}
\caption{Adapted CRISP data mining cycle. CRISP is characterised by
its independence from the application domain and the algorithms
used. This makes it suitable as base data mining cycle where
metacognitive aspects (also independent from domain and algorithms)
are included.}
\label{cycle}
\end{figure}

The last two steps, integration and evaluation, are implemented by
augmenting the data mining life cycle to support a live integration of
obtained models.  We call this additional step the {\it (automatic)
  operationalisation} of learned meta models. Figure \ref{cycle}
illustrates the Cross Industry Standard Process for Data Mining
cycle\footnote{See {\it http://www.crisp-dm.org}.} and  includes our
augmentation.  In the {\it modelling phase}, various modelling
techniques are selected and applied.  The modelling phase is finished
when one or more models, which appear to be of high quality at least
from a data analysis perspective, have been built.  These models then
need to be evaluated before their deployment.  In the {\it evaluation
  phase} we use the models to review the model building process. This
{\it evaluation} is done by running the system on unseen supervised
data or by reinforcement learning experiments.  Finally, at the end of
the evaluation stage, a decision has to be reached as to whether to use
the data mining results obtained.  Then a new model is deployed and
used in the domain or business units.

The CRISP cycle closes with the evaluation of the deployed system in the
real application context ({\it domain/business understanding}), whether 
it performs well, or not.  In fact, this is a kind of
metacognitive process conducted by the domain experts. The
introspective mechanism represents a new phase between evaluation and
(human) domain/business understanding.  It automatically optimises the
behaviour of the deployed system and provides hints for human
understanding by generating transparent metamodels of the system's performance,
for example, introspective association rules and decision trees.  The
cycle now includes the additional step  {\it (automatic)
  operationalisation} before it closes.

Our aim to integrate the introspective mechanism in order to extend
the data mining cycle by a new phase where system introspection is
integrated, resulted in a new step of the data mining life cycle,
i.e., {\it (automatic) operationalisation}.  The introspective models
are directly used in conjunction with the former decision making
models for action taking.  As a result, the augmentation of the CRISP
cycle represents a tractable metacognitive model creation and
verification process. In subsequent applications of the augmented
CRISP cycles, the introspective models can be combined with the models
of the former CRISP process.

It is important to note that empirical machine learning models are
pattern patching systems; we expect the behaviour to be improved by
drawing an analogy to a past experience which materialises as patterns
to be mined. These patterns do not necessarily follow logical rules in
terms of a higher order logic---but instead, they should follow at
least the causal implications of a propositional logic which helps to
implement reactivity based on learned causality. All patterns to be
mined can be regarded as {\it introspective reports} on the
application or business domain.

\section{Conclusion}

The question we investigated was about the scope and usefulness of a
metacognitive model. In order to develop a computational introspective
model, empirical machine learning models can be investigated. This
should augment cognitive capabilities of adaptable AI systems,
especially in the reasoning phase before action taking, which we
believe requires to a great extent metacognitive instead of cognitive
capabilities.

Similar methodology in computation has received great attention for
uncertainty handling, control in decentralised systems, scheduling for
planning in real-time, and meta-level reasoning in general
\cite{metacognition2005}.  Applications are to be found in the
contexts of large-scale natural language processing architectures for
texts (e.g., UIMA \cite{uima04}), and dialogical interactions with the
Semantic Web (e.g., SmartWeb \cite{sonntagetal07} integrating
extensive ontological groundwork \cite{dolcesumo07} for
self-representation of an information state to be included into a
metacognitive model). The metacognitive control and augmented Data
Mining Cycle proposed here will be integrated into a new
situation-aware dialogue shell for the Semantic Access to Media and
Services in the near future---to handle, fore and foremost, the access
to dynamic, heterogeneous information structures.

\subsubsection{Acknowledgements.}
This research has been supported in part by the THESEUS Program in the
Core Technology Cluster WP4 {\it Situation Aware Dialogue Shell for
  the Semantic Access to Media and Services}, which is funded by the
German Federal Ministry of Economics and Technology under the grant
number 01MQ07016. The responsibility for this publication lies with
the author.

\bibliographystyle{splncs}
%\scriptsize{
\bibliography{diss} 
%}
\end{document}